\documentclass[journal]{IEEEtran}

\usepackage{subfigure}
\usepackage{palatino,epsfig,latexsym,cite,graphicx,amsmath,amssymb,amsfonts,multirow,booktabs,color,soul}
\usepackage{algorithmic}
\usepackage{float}
\usepackage{threeparttable}
\usepackage{amsmath}
\usepackage[linesnumbered,ruled,vlined]{algorithm2e}
\def\MR2{\multirow{2}[2]{*}}

\definecolor{hl}{rgb}{0.75,0.75,0.75}
\newcommand{\XG}[1]{\textcolor[rgb]{0.00,0.00,0.00}{#1}}

\sethlcolor{hl}
\usepackage{makecell}
\usepackage[caption=false,font=footnotesize]{subfig}

\begin{document}

\title{Deep Reinforcement Learning-Assisted Automated Operator Portfolio for Constrained Multi-objective Optimization
}
\author{
        Shuai Shao,
        Ye Tian,~\IEEEmembership{Senior Member,~IEEE,}
        Shangshang Yang,
        and
        Xingyi Zhang,~\IEEEmembership{Fellow,~IEEE}
\vspace*{\fill}
\begin{center}
	\footnotesize
	© 2026 IEEE. Personal use of this material is permitted. Permission from IEEE must be obtained for all other uses.
	
	This article has been accepted for publication in \textit{IEEE Transactions on Emerging Topics in Computational Intelligence}.
\end{center}
\thanks{S. Shao, Y. Tian, S. Yang, and X. Zhang are with the School of Computer Science and Technology, Anhui University, Hefei 230601, China (email: freshshao@gmail.com; field910921@gmail.com; yangshang0308@gmail.com; xyzhanghust@gmail.com).}
}

\markboth{IEEE Transactions on Emerging Topics in Computational Intelligence,~Vol.~, No.~, month~year}
{\MakeLowercase{Tian \textit{et al.}}: No title}

\IEEEpubid{0000--0000/00\$00.00~\copyright~0000 IEEE}

\maketitle

\begin{abstract}
Constrained multi-objective optimization problems (CMOPs) are of great significance in the context of practical applications, ranging from scientific to engineering domains. Most existing constrained multi-objective evolutionary algorithms (CMOEAs) usually employ fixed operators all the time, which exhibit poor versatility in handling various CMOPs. Therefore, some recent studies have focused on adaptively selecting the best operators for the current population states during the search process. The evolutionary algorithms proposed in these studies learn the value of each operator and recommend the operator with the highest value for the current population, resulting in only a single operator being recommended at each generation, which can potentially lead to local optima and inefficient utilization of function evaluations. To address the dilemma in operator adaptation, this paper proposes a reinforcement learning-based automated operator portfolio approach to learn an allocation scheme of operators at each generation. This approach considers the optimization-related and constraint-related features of the current population as states, the overall improvement in population convergence and diversity as rewards, and different operator portfolios as actions. By utilizing deep neural networks to establish a mapping model between the population states and the expected cumulative rewards, the proposed approach determines the optimal operator portfolio during the evolutionary process. By embedding the proposed approach into existing CMOEAs, a deep reinforcement learning-assisted automated operator portfolio based evolutionary algorithm for solving CMOPs, abbreviated as CMOEA-AOP, is developed. Empirical studies on 33 benchmark problems demonstrate that the proposed algorithm significantly enhances the performance of CMOEAs and exhibits more stable performance across different CMOPs.
\end{abstract}

\begin{IEEEkeywords}
Multi-objective optimization, constrained optimization, evolutionary computation, deep reinforcement learning, operator portfolio
\end{IEEEkeywords}


\section{Introduction}

\IEEEPARstart{M}{any} real-world optimization problems involve multiple objectives and constraints, which are referred to as constrained multi-objective optimization problems (CMOPs). For example, the vehicle routing problem with time window aims to minimize the total distance of routes and the number of vehicles, while the solution must ensure the delivery time is within the time window specified by each customer \cite{duan2021robust}. The robot gripper optimization problem aims to minimize the fluctuations in grasping force and force transmission ratio, and its solutions are constrained by six constraints related to geometry and force considerations \cite{datta2015analysis}. The multi-objective testing resource portfolio problem aims to minimize three objectives of reliability, cost, and time, while a variety of time and reliability constraints need to be met~\cite{su2021enhanced}. 

\IEEEpubidadjcol

Due to the constraints of the above problems altering the shape of the search space and creating infeasible regions, CMOPs pose greater challenges compared with their unconstrained counterparts \cite{qiao2024benchmark, yang2024dynamic}. The ultimate goal in solving CMOPs is to obtain a set of solutions with good convergence and diversity that reside in the feasible regions. Over the past twenty years, multi-objective evolutionary algorithms (MOEAs) have been proven effective in solving CMOPs, as they can preferentially handle constraints during optimization to obtain a set of solutions that compromise between constraints and optimization \cite{li2024decoupling}. A constrained multi-objective evolutionary algorithm (CMOEA) typically comprises two parts \cite{qiao2024benchmark}: MOEA and constraint-handling techniques (CHTs), which are utilized for optimizing objectives and tackling constraints, respectively. CHTs are often embedded into the environmental selection process or mating selection process of MOEAs to select high-quality solutions. Although a large number of CMOEAs have been developed by combining different MOEAs and CHTs, these CMOEAs usually utilize fixed variation operators to generate offspring solutions. However, according to the no free lunch theorem, there is no single variation operator that outperforms all others across all optimization problems. As a result, the majority of the existing CMOEAs struggle with generalization due to the limitations of singular search paradigms.

\XG{To be specific, in evolutionary algorithms, variation operators mainly include crossover operators \cite{deb1995simulated}, mutation operators \cite{katoch2021review}, swarm intelligence operators \cite{shami2022particle}, probability models \cite{iruthayarajan2010covariance}, and generative models \cite{tian2020solving}.} \XG{Most variation operators include a crossover operator to propagate the beneficial genes of the parent individuals and a mutation operator to enhance exploration capability. Specifically, the crossover operator \cite{xue2021adaptive, shao2023non} generates new offspring individuals by exchanging and combining the chromosomes of parent individuals, accelerating the evolution towards the Pareto front, while the mutation operator \cite{alhijawi2023genetic, tian2024multi} enhances population diversity by modifying individual genomes, preventing the evolutionary algorithm from getting stuck in local optima.} \XG{There are also some operators that do not strictly follow crossover and mutation, but instead directly generate offspring using methods such as neural networks \cite{tian2024neural} and particle swarm optimization \cite{shami2022particle}.} The selection and tuning of these operators are crucial for the performance and different variation operators exhibit quite different search paradigms \cite{piotrowski2023particle}. For example, differential evolution \cite{chakraborty2023differential} mutates each solution based on the differences between other solutions, making it adept at handling complex variable interactions. Particle swarm optimization \cite{yang2023random} updates solutions by learning from local and global best solutions, enabling rapid convergence in high-dimensional search spaces. Simulated binary crossover \cite{deb1995simulated} simulates the idea of binary crossover to achieve efficient search in continuous decision space, particularly adept at handling optimization problems with multimodal features.

Due to the population being distributed across different function landscape regions during the iteration process of optimization algorithms \cite{yan2021leader, qiao2023evolutionary}, variation operators should be carefully selected when solving optimization problems. Thus, an operator should be selected from some candidates through empirical comparisons \cite{li2013adaptive}. However, such a trial-and-error process is impractical due to countless optimization problems. To address this issue, some works adaptively implement automated operator selection through deep reinforcement learning \cite{tian2022deep, ming2024constrained}. Deep reinforcement learning uses deep neural networks to learn the mapping relationship between complex population states and candidate operators, enabling adaptive selection of operators. This method automatically adjusts strategies through an interactive learning process with evolutionary algorithms and possesses excellent generalization capabilities, allowing it to adapt to unseen optimization problems.

While the adaptive selection of operators has garnered increasing attention in the multi-objective optimization community, unfortunately, few studies have focused on constrained multi-objective optimization. This is because constrained optimization requires the simultaneous consideration of both constraints and objectives. More importantly, most existing works for operator selection can only recommend a single operator for the current population, leading to an excessive bias towards either exploration or exploitation during a single iteration process \cite{ming2022competitive, ming2024even}. Therefore, to reduce the probability of falling into local optima and achieve a better balance between optimization and constraints \cite{afshari2019constrained, ming2021simple}, this paper employs reinforcement learning to develop an operator portfolio approach for solving CMOPs, which determines the probability of using each operator rather than selecting a single one at each generation. The main contributions are reflected in the following three aspects:

\begin{enumerate}
	\item We propose a deep reinforcement learning-assisted automated operator portfolio approach. By treating the optimization-related and constraint-related features of the current population as states, operator portfolio schemes as actions, and the overall improvement of population convergence and diversity as rewards, the population evolution is regarded as a Markov decision process. A reinforcement learning agent utilizes deep neural networks as its actor and critic networks, where the actor network outputs a policy of the current operator portfolio scheme, and the critic network evaluates the cumulative rewards of the actor's policy and guides its updates. The agent learns future rewards rather than past rewards to recommend the optimal operator portfolio for the current population states, thereby aiming to generate promising offspring solutions in subsequent generations.
	
	\item A deep reinforcement learning-assisted automated operator portfolio based evolutionary algorithm for solving constrained multi-objective optimization, abbreviated as CMOEA-AOP, is developed. In the proposed algorithm, the agent periodically updates the weights of actor and critic networks. By adopting different operator portfolio schemes as candidate actions, the proposed algorithm is less likely to fall into local optima. To the best of our knowledge, this is the first work that uses reinforcement learning to learn the optimal operator portfolio rather than recommending a single operator for evolutionary algorithms.
	
	\item To evaluate the performance of the proposed algorithm, comprehensive empirical experiments have been conducted on a variety of CMOPs in comparison with several state-of-the-art CMOEAs for solving CMOPs. The experimental results demonstrate that the proposed algorithm has significant advantages in terms of the convergence, diversity, and feasibility on different CMOPs, especially on CMOPs with complex feasible regions.
	
\end{enumerate}

The remainder of this paper is structured as follows: Section II offers a brief overview of fundamental concepts related to CMOPs and existing adaptive operator selection approaches, followed by a discussion on the motivation for this work. In Section III, we present a detailed explanation of the proposed MOEA. Section IV showcases our experimental results and analysis on CMOPs. Finally, we draw conclusions and discuss future work in Section V.

\section{Related Work and Motivation}

\subsection{Constrained Multi-objective Optimization}

In general, a CMOP involves optimizing multiple objective functions while satisfying certain constraints, which can be mathematically defined as 
\begin{equation}
	\begin{aligned}
		\text{Minimize\quad}& \mathbf{F}(\mathbf{x})=(f_1(\mathbf{x}),\ldots,f_M(\mathbf{x}))  \\
		\text{subject to\quad}& \mathbf{x}\in\Omega   \\
		&g_i(\mathbf{x})\leq0,\quad i=1,\ldots,p \\
		& h_j(\mathbf{x})=0,\quad j=1,\ldots,q 
	\end{aligned}
\ ,
\end{equation}
where $\mathbf{x}=(x_1,x_2,\ldots,x_D)$ represents a $D$-dimensional decision vector, $f_1(\mathbf{x}),\ldots,f_M(\mathbf{x})$ indicates $M$ objective functions, and $g_i(\mathbf{x}),h_j(\mathbf{x})$ indicate the inequality constraints and equality constraints, respectively. The constraint function values can be calculated:
\begin{equation}
	CV(\mathbf{x}) = \sum_{i=1}^p gv_i(\mathbf{x}) + \sum_{j=1}^q hv_j(\mathbf{x})\ ,
\end{equation}
where $gv_i(\mathbf{x}) = \max(0, g_i(\mathbf{x}))$, $hv_j(\mathbf{x}) = \max(0, | h_j(\mathbf{x}) | - \delta)$ ($\delta$ is a very small positive number to relax equality constraints). 
If $CV(\mathbf{x})=0$, $\mathbf{x}$ represents a feasible solution; otherwise, it is an infeasible solution. The mapping vectors in the objective space of Pareto optimal solutions that satisfy constraints are called the constrained Pareto front. It is worth noting that in the absence of constraints, the Pareto front is called the unconstrained Pareto front.

When solving CMOPs, the aim is to identify and search for a set of feasible Pareto optimal solutions with good convergence and diversity. To achieve this goal, some customized evolutionary algorithms for constrained multi-objective optimization, known as CMOEAs, have been recently proposed, which can be grouped into four categories \cite{tian2021balancing, qiao2024benchmark, wu2024surrogate}. The first category covers CMOEAs that focus on CHTs \cite{deb2002fast, yang2019multi, liu2022multiobjective}. These CHTs influence the selection of solutions by expressing different preferences for constraints and objective values and can be easily integrated into general evolutionary algorithms. However, due to parameter settings or preference settings, these CHTs can not effectively address different CMOPs. The second category addresses CMOPs based on the multi-stage mechanism \cite{fan2019push, liang2022utilizing}. To be specific, constraints and optimizations are handled separately in different stages of the evolutionary process, resulting in rapid convergence. However, specifying the switching conditions between multiple stages is challenging. The third category achieves a dynamic balance between objectives and constraints by employing different strategies in different populations \cite{ming2022tri, qiao2024dual}. For instance, CCMO \cite{tian2020coevolutionary} evolves one population to address the original CMOP and another population to tackle auxiliary problems derived from the original population, thus achieving the separation of constraints and optimizations. The fourth category addresses CMOPs based on the multi-task mechanism \cite{qiao2022evolutionary, ming2022constrained, qiao2023evolutionary}. This type of CMOEAs also utilize a dual-population mechanism but capture migration information during the evolutionary process to facilitate better interaction between the two populations.

While constrained multi-objective optimization has been studied for twenty years, the above CMOEAs still have some limitations \cite{liang2022survey, qiao2024dual}. To be more specific, the above CMOEAs balance constraints and optimizations through multiple solution selection strategies. By contrast, a single operator is usually employed for offspring generation, which may struggle to address different types of CMOPs. To enhance the algorithm's generality, some adaptive operator selection approaches \cite{tian2022deep, zuo2023process} have been proposed to achieve optimal operator selection during the evolutionary process, which are reviewed in the next subsection.

\subsection{Existing Operator Adaptation Approaches}

Adaptive operator selection  \cite{hitomi2016classification, santiago2019novel} is a class of approaches dedicated to enhancing the generality of MOEAs and other metaheuristics. Adaptive operator selection acts as an advanced controller selects operators during the evolutionary process to explore the decision space. The two main components of the adaptive operator selection are the credit assignment policy, which defines how operators are rewarded based on their impact during the search process, and the operator selection policy, which uses these rewards to determine the next operator to be used \cite{hitomi2016classification}. When solving MOPs, the credit assignment policy often relies on the improvement of population convergence and diversity as the rewards for credit portfolio. Operator selection policy makes varying degrees of greedy selection based on the result of credit portfolio. For example, probability matching \cite{goldberg1990probability} selects operators based on their credit, with each operator's selection probability proportional to its credit.  However, probability matching may struggle when there are many mediocre operators and only a few high-performing ones. In contrast, adaptive pursuit \cite{thierens2007adaptive} addresses this by greedily selecting the highest-credit operator and asymptotically pursuing it while assigning others the minimum probability. Moreover, various selection approaches have been employed for operator selection, including the choice functions \cite{gonccalves2015moea}, fuzzy inference system \cite{santiago2019novel}, dynamic island models \cite{candan2012dynamic}.

Although selecting operators with many rewards is advantageous, occasionally exploring poorly performing operators can also be beneficial, as they may start generating high-quality solutions as the population evolves \cite{hitomi2016classification, tian2022deep}. To be specific, the above-mentioned adaptive operator selection approach learns the rewards of operators through historical experience, but operators that perform well in the early generations may not perform well in future generations. To mitigate this dilemma, some work utilizes deep reinforcement learning to learn the future rewards of candidate operators \cite{bai2023evolutionary, majid2023deep, li2024bridging}. Specifically, reinforcement learning is a machine learning approach that learns how to make optimal decisions through interactions with the environment. During the interaction process, the agent gradually optimizes its behavior policy to maximize the accumulated reward by observing the current population states, taking actions, and receiving rewards. The process of action selection by the agent involves using value functions or policy functions to evaluate the pros and cons of each action, thereby enabling the selection of the best operator for different population states. Using deep reinforcement learning to achieve adaptive operator selection is a promising approach as it learns from future rewards rather than past ones \cite{mankowitz2023faster, zhu2023transfer}.

The pioneer in achieving adaptive operator selection through reinforcement learning is MOEA/D-DQN, proposed by Tian \emph{et al.} \cite{tian2022deep}, which employs deep reinforcement learning networks during the optimization process to learn the relationship between decision variables and the \emph{Q} values of candidate operators, i.e., the cumulative fitness improvement. This work demonstrates the promising potential of reinforcement learning in aiding evolutionary algorithms to solve MOPs. Due to MOEA/D-DQN treating only the decision variables of a single solution as states, it struggles to accurately select suitable operators for different types of problems. Therefore, AdaW-DDQN \cite{yin2024adaptive} extracts features from the population, incorporating the improvement of individual fitness and the average Euclidean distance of the population in decision space and objective space as states. There are also few works through reinforcement learning to select operators for CMOPs. \XG{Ming \emph{et al.} \cite{ming2024constrained} design specialized states for constrained multi-objective optimization to better adaptively select operators for balancing constraints and optimization.} 

\subsection{Motivation of This Work}

\begin{figure}[!t]
	\centering
	\subfigure{\includegraphics[width=0.85\linewidth]{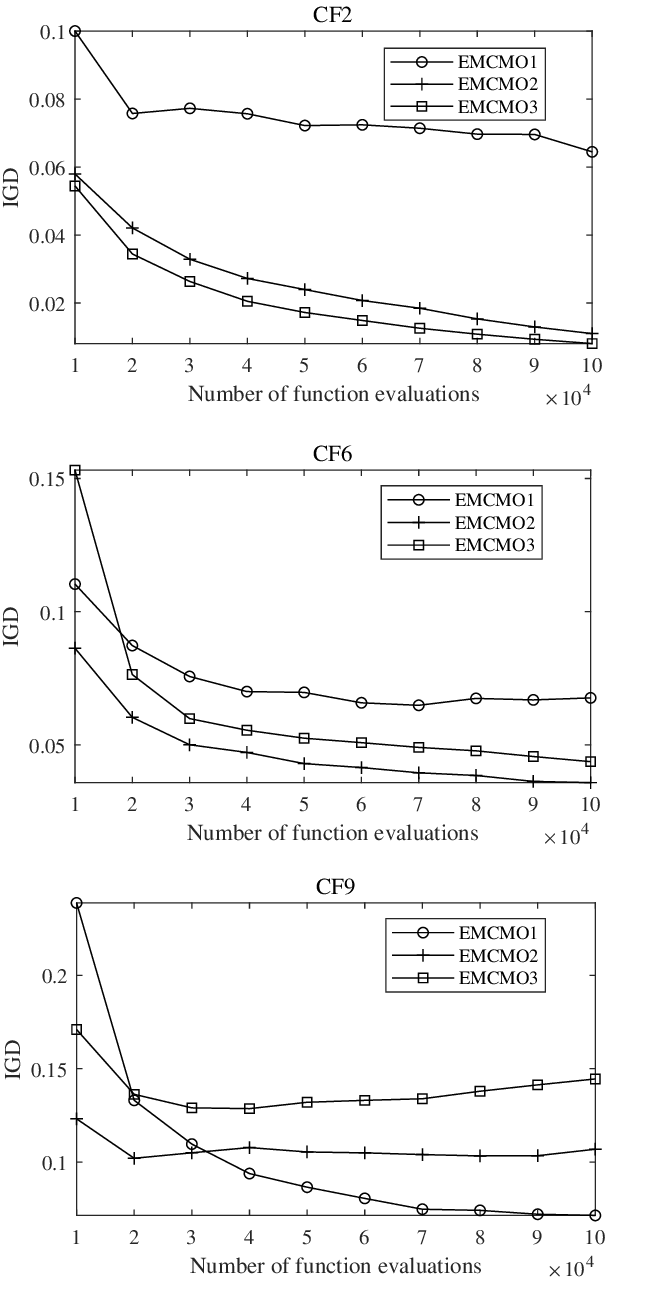}}
	\caption{Convergence profiles of IGD values obtained by EMCMO1 (using only genetic operators), EMCMO2 (using only differential evolution operators), EMCMO3 (using both genetic and differential evolution operators) on CF2, CF6, and CF9.}
	\XG{\label{fig:EMCMODifferentOP}}
\end{figure}

Fig.~\ref{fig:EMCMODifferentOP} displays the convergence profiles of three variants of a state-of-the-art algorithm EMCMO \cite{qiao2022evolutionary}, using different operator portfolio schemes. Specifically, EMCMO1 and EMCMO2 use genetic operators and differential evolution operators, respectively, while EMCMO3 generates half of the offspring solutions using genetic operators and the other half using differential evolution operators. It is evident that the best operator portfolio scheme varies for different CMOPs, underscoring the importance of using an appropriate operator portfolio scheme for solving various CMOPs. Unfortunately, most existing CMOEAs employ a fixed operator for all CMOPs, resulting in poor generalization. Furthermore, existing approaches for reinforcement learning based adaptive operator selection can only recommend a single operator for the current evolutionary states \cite{tian2022deep, yin2024adaptive, ming2024constrained}, leading to evolutionary algorithms being confined to a fixed paradigm in decision space. This limitation is particularly pronounced in constrained multi-objective optimization, where search must simultaneously consider both constraints and optimizations. Consequently, existing adaptive operator selection approaches struggle to escape local optima or reach feasible regions due to the singular search paradigm during a single iteration. 

\XG{It is worth noting that some CMOEAs construct penalty terms through dynamic or adaptive approaches to influence the objective function, thereby achieving a preference for the feasible solutions. For example, Maldonado and Zapotecas-Martínez \cite{maldonado2021dynamic} propose a dynamic penalty function approach, in which the parameters gradually change with the generation number and are embedded into the MOEA/D to solve CMOPs. Long \emph{et al.} \cite{long2023constrained} propose a fitness function that prioritizes infeasible solutions near the constraint boundary and dynamically adjusts the trade-off between constraints and objectives based on the proportion of feasible solutions, accelerating convergence to constrained Pareto optimal solutions. ICMOEA/D \cite{zhou2024adaptive} adaptively determines the penalty coefficients using the mean of the objective values and the average violation value for each constraint in the solutions, thereby achieving a better balance between optimization and constraints. Although these CMOEAs have the capability for adaptive adjustments to CMOPs of varying difficulty, they focus only on solution selection while still employing fixed operators for solution generation.}

\begin{figure*}[!t]
	\centering
	\subfigure{\includegraphics[width=0.75\linewidth]{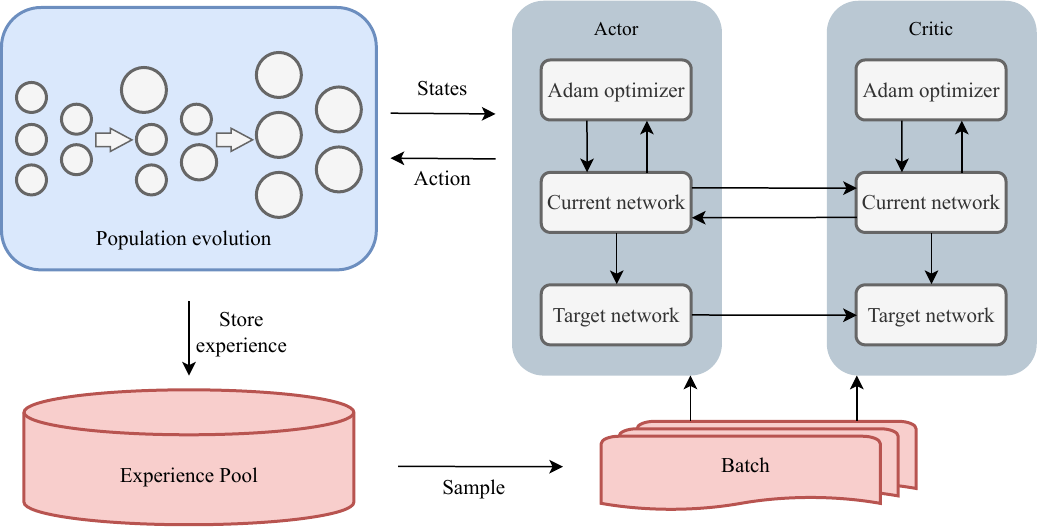}}\\
	\caption{\XG{Illustration of the proposed CMOEA-AOP with deep reinforcement learning based automated operator portfolio, where a DDPG agent consisting of an actor network and a critic network is employed to recommend the optimal operator portfolio scheme for CMOEAs. Here ``States'' represent the optimization-related and constraint-related features of the current population, and ``Action'' represents the optimal operator portfolio scheme recommended by the deep reinforcement learning agent based on the current evolutionary states.}}
	\label{fig:framework}
\end{figure*}

In this paper, we propose an evolutionary algorithm that integrates an automated operator portfolio approach based on deep reinforcement learning. \XG{Deep reinforcement learning is a powerful machine learning paradigm that learns optimal policies through interactions between an agent and the environment. Compared to other decision-making approaches, deep reinforcement learning focuses on maximizing long-term rewards rather than immediate gains, making it less likely to get trapped in local optima  \cite{bai2023evolutionary, ming2024constrained}.} \XG{Considering the excellent performance of Deep Deterministic Policy Gradient (DDPG) \cite{sumiea2024deep} in complex environments, }\XG{we use DDPG  to build a reinforcement learning agent to learn the mapping models between optimization-related features, constraint-related features, and operator portfolio schemes, enabling the best recommendation of all operators based on the current population characteristics.} The operator portfolio schemes recommended by the reinforcement learning agent result in offspring solutions generated with different proportions through various operators during a single iteration. In contrast to existing approaches that can only learn and recommend a single operator in a single iteration, the proposed algorithm automatically recommends the optimal operator portfolio scheme based on the current population's characteristics. The proposed algorithm facilitates faster convergence towards the constrained Pareto optimal solutions while avoiding local optima, thereby achieving better generalization performance. The next section will provide detailed insights into the proposed algorithm.

\section{The Proposed Algorithm}

This section presents a constrained multi-objective evolutionary algorithm with reinforcement learning-assisted automated operator portfolio, referred to as CMOEA-AOP. Firstly, in the early stages of evolution, different operator portfolio schemes are evaluated for rewards under different population states. Then, based on the collected training samples, a mapping model (i.e., reinforcement learning agent) is established between population states and operator portfolio schemes using the reinforcement learning agent. Finally, based on the established mapping model and the current population states, the agent recommends subsequent operator portfolio schemes and updates the model through future rewards. In the following sections, we will provide the proposed algorithm framework, as well as the specially designed states, actions, and rewards for training the agent tailored for constrained multi-objective optimization. Finally, we present details on the training and application of the reinforcement learning agent.

\subsection{Framework of CMOEA-AOP}

\begin{algorithm}[!t]
	\caption{Main procedure of CMOEA-AOP}
	\label{alg:main}
	\SetKwComment{Comment}{//}{}
	\KwIn{$N$ (population size), $FE_{max}$ (maximum number of evaluations)}
	\KwOut{$P$ (final population)}
	$Agent \leftarrow$ Initialize the reinforcement learning agent\;
	$EP \leftarrow \emptyset$; \Comment{Experience pool}
	$P\leftarrow$; \Comment{Initialize the population using the employed CMOEA}
	$FE\leftarrow |P|$; \Comment{Number of consumed evaluations}
	\While{$FE\leq FE_{max}$}
	{
		$Action \leftarrow$ Operator portfolio scheme for the current population $P$ is recommended by $Agent$\;
		$O \leftarrow$ Generate offspring solutions by adopting an existing CMOEA using the operator portfolio scheme $Action$\;
		\XG{$P\leftarrow$ Select $N$ solutions for the next generation by adopting an existing CMOEA\;}
		$EP \leftarrow $ Determine the reward and states, and insert the record to EP\;
		$Agent \leftarrow Training(Agent,EP)$; \Comment{Algorithm \ref{alg:train}}
		$FE \leftarrow$ Update the consumed evaluations\;
	}
	\textbf{return} $P$\;
\end{algorithm}

Algorithm \ref{alg:main} and Fig. \ref{fig:framework} illustrate the main procedure of CMOEA-AOP. The proposed algorithm takes as input the population size $N$ and the maximum number of evaluations $FE_{max}$, and outputs the final population $P$. It begins by randomly initializing a reinforcement learning agent $Agent$ and an empty experience memory pool $EP$. Then, it initializes the population $P$ using an employed CMOEA, and the number of consumed evaluations $FE$ to the size of the population $|P|$. At each generation, while the number of consumed evaluations is less than or equal to the maximum number of evaluations, the following steps are performed: first, the reinforcement learning agent is used to recommend the operator portfolio scheme according to the current population states. Then, generate offspring solutions using the employed CMOEA and the current action $action$, and select $N$ solutions for the next generation using the employed CMOEA. Next, determine the reward and state, and insert the record into the experience memory pool $EP$. After that, train the reinforcement learning agent $Agent$ using the experience memory pool $EP$ to update its parameters. Finally, update the number of consumed evaluations $FE$. Once the main loop terminates, the final population $P$ is returned. 

It can be seen that the main contribution of the proposed algorithm lies in using the reinforcement learning agent to recommend suitable operators for generating offspring according the current population states. For other constraint handling mechanisms, most existing CMOEAs such as CCMO \cite{tian2020coevolutionary} and EMCMO \cite{qiao2022evolutionary} can be employed. \XG{In this paper, we employ EMCMO for empirical studies.} The key to training the reinforcement learning agent lies in how to define the states, actions, and rewards. In the next subsection, we provide a detailed explanation.

\subsection{Automated Operation Portfolio}

\textbf{Actions.} In this work, we use the genetic operators (i.e., simulated binary crossover) and two differential evolution operators (i.e., DE/rand/1 and DE/best/1) as candidate operators for instantiating the reinforcement learning agent. These are the most commonly used operators in existing CMOEAs, where genetic operators are well-suited for handling multimodal problems with strong convergence capabilities, and differential evolution operators are proficient in handling dependencies and adept at exploring the decision space. In addition, DE/rand/1 and DE/best/1 focus on maintaining population diversity and convergence, respectively. The reinforcement learning agent will recommend the proportion of each operator to be used for the current population states. It can be seen that in our work, the action is not a single operator, but a proportion of the use of candidate operators. The reasons for this are two-fold:

Firstly, population in the evolutionary process often occupies regions of the function landscape that exhibit multiple complex relationships, such as multimodality and linkage. Using only one operator in a single iteration leads to a singular search paradigm, making evolutionary algorithms prone to getting trapped in local optima regions and hindering further convergence in later stages.

Secondly, when using a single operator as a candidate action for reinforcement learning, the stochastic nature of evolutionary algorithms can lead to instances where a single operator accidentally receives a large reward, which may confuse the reinforcement learning agent and result in an excessive recommendation of that particular operator. In our work, the designed actions mostly involve all operators simultaneously, with only varying proportions. This ensures that even if reinforcement learning tends to over-recommend, each operator is still used in a certain proportion in each iteration.

\textbf{States.}  We define the state of the population by considering its performance in terms of convergence $con$, diversity $div$, feasibility $fea$, and evolutionary stage $stage$. The convergence of the population is assessed by the average value of each objective function, indicating the degree of approximation to the Pareto optimal solutions. Specifically, each element of $con$ is formulated as
\begin{equation}
	con_i=\frac{\sum_{\mathbf{x}\in\mathcal{P}}f_i(\mathbf{x})}N\ ,
\end{equation}
where $f_i(\mathbf{x})$ is the $i$-th objective function value. Therefore, $con = (con_1, con_2, \dots, con_m)$ ($m$ is the number of objective functions). When the evolutionary algorithm performs well on a specific objective function, the corresponding element in $con$ will have a smaller value. Similarly, if a specific objective function is difficult to optimize during the evolution process, the corresponding element in $con$ will have a larger value. In this way, the convergence of multiple objective functions can be reflected simultaneously.

The degree of dispersion of the population on each objective function is used to assess the diversity of the population. Specifically, each element of $div$ is formulated as
\begin{equation}
	div_i=\frac{(\sum_{\mathbf{x}\in\mathcal{P}}f_i(\mathbf{x}) - con_i)^2}N\ ,
\end{equation}
where $con_i$ is the average value of $i$-th objective function. When the solutions of the population are clustered together, it indicates poor diversity in the population, and each element of $div$ will have a smaller value. Conversely, when the solutions of the population are spread out over a wider range, the diversity of the population is better. In this case, each element of $div$ will have a larger value.

The average CV value of the population is used to estimate feasibility, that is, the extent to which the population satisfies constraints. This indirectly reflects whether the population is within the feasible region. Specifically, each element of $fea$ is formulated as
\begin{equation}
	fea=\frac{\sum_{\mathbf{x}\in\mathcal{P}}cv(\mathbf{x})}N\ ,
\end{equation}
where $cv(\mathbf{x})$ is the CV of $\mathbf{x}$. When the value of $fea$ is 0, it indicates that all solutions are entirely within the feasible region. Conversely, when 
$fea$ is relatively large, it implies that some solutions are outside the feasible region. Using the cumulative constraint violation of each solution as part of the population states can effectively reflect the current population's constraint handling situation.

In addition, we also consider the current evolutionary stage $\lambda$, i.e., the proportion of evaluations consumed compared to the maximum evaluations, as a state. As a result, a state $s=(con, div, fea, \lambda)$ consists of the convergence, diversity, feasibility, and evolutionary stage of the population. This way, the optimization-related and constraint-related features of the population can be extracted, thereby guiding the reinforcement learning agent to recommend the optimal operator portfolio scheme for the current population.

\textbf{Rewards.} In multi-objective optimization, the hypervolume metric (HV) \cite{zitzler1999multiobjective} serves as a comprehensive indicator for evaluating the performance of solution sets. It quantifies the volume enclosed by the solution set obtained by the algorithm, providing insights into both convergence and diversity aspects of the population. Therefore, the improvement in the hypervolume value of the current population relative to its predecessor is utilized as a reward signal for the automated operator portfolio approach. This reward mechanism enables the assessment of how effectively the adopted operator portfolio scheme enhances both convergence and diversity within the algorithm.

Based on the actions, states, and rewards defined above, a record of experience memory pool $EP$ is formed as $m = (s_t, a_t, r_t, s_{t+1})$. Through continuous interaction between the agent and the environment, more training samples are stored in the experience pool $EP$, further enhancing the decision-making capability of the agent. In the next subsection, we will provide details on training and using the reinforcement learning agent.

\subsection{Agent Training}

\begin{algorithm}[!t]
	\caption{$Training(Agent,EP)$}
	\label{alg:train}
	\SetKwComment{Comment}{//}{}
	\KwIn{$Agent$ (the reinforcement learning agent), $EP$ (experience memory pool), $BS$ (Batch size)}
	\If{$|EP| < BS$}
	{
		No training\;
	}
	\Else
	{
		$Batch \leftarrow$ Sample random mini-batch of transitions from $EP$\;
		$s_t \leftarrow$ The state of the population before performing the action, $t\in Batch$\;
		$a_t \leftarrow$ Recommended operator portfolio scheme for population state $s_t$, $t\in Batch$\;
		$r_t \leftarrow$ Reward after performing the action, $t\in Batch$\;
		$s_{t+1} \leftarrow$ The state of the population after performing the action, $t\in Batch$\;
		Update critic network parameters by minimizing the loss using Eq. (7)\;
		Update actor network parameters $\theta_{\mu}$ using policy gradient ascent using Eq. (9)\;
		Update target networks parameters $\theta_{\mu'}$ and $\theta_q'$ using Eq. (10)\;
		
	}
	\KwOut{$Agent$ (the reinforcement learning agent)}
	
	\textbf{return} $Agent$\;
\end{algorithm}

Algorithm \ref{alg:train} delineates the training procedure of the reinforcement learning agent. In this work, we use DDPG to learn the mapping relationship between population states and operator portfolio schemes. DDPG is a reinforcement learning algorithm designed for continuous action spaces. It combines deep learning with traditional deterministic policy gradient methods, enabling effective policy optimization in complex, high-dimensional environments. To be specific, DDPG maintains four neural networks:
\begin{itemize}
	\item Actor Network: $\mu(s|\theta^\mu)$
	\item Critic Network: $Q(s, a|\theta^Q)$
	\item Target Actor Network: $\mu'(s|\theta^{\mu'})$
	\item Target Critic Network: $Q'(s, a|\theta^{Q'})$
\end{itemize}
The actor network aims to learn a deterministic policy function that takes the state as input and outputs a continuous action, i.e., the operator portfolio schemes. The actor network is typically modeled using deep neural networks, which can include multiple hidden layers. The activation function used in the final output layer is softmax to ensure the output falls within the action space range. The critic network aims to estimate the policy performance of the actor network. Given a state and its corresponding action, the critic network outputs a value function $Q(s, a)$, which represents the value of that action in that population states. Similar to the actor network, the critic network is also modeled using deep neural networks, where the inputs are the state and action, and the output is the corresponding value. The training objective of the critic network is to minimize the mean squared error between the predicted value and the target value. In addition, the two target networks are delayed copies of the original networks and help stabilize training. During training, mini-batches of experiences are sampled from the experience pool $EP$ to break the correlation between consecutive samples and improve the efficiency of learning. The critic network is updated by minimizing the loss function:
\begin{equation}
	L(\theta^Q) = \mathbb{E}_{(s_t, a_t, r_t, s_{t+1}) \sim R} \left[ \left( Q(s_t, a_t | \theta^Q) - y_t \right)^2 \right]\ ,
\end{equation}
where the target value $y_t$ is defined as
\begin{equation}
	y_t = r_t + \gamma Q'(s_{t+1}, \mu'(s_{t+1}|\theta^{\mu'})|\theta^{Q'})\ .
\end{equation}
Here, $\gamma$ is the discount factor. The actor network is updated by maximizing the expected return using the policy gradient:
\begin{equation}
	\nabla_{\theta^\mu} J \approx \mathbb{E}_{s_t \sim R} \left[ \nabla_a Q(s_t, a | \theta^Q) \big|_{a=\mu(s_t)} \nabla_{\theta^\mu} \mu(s_t | \theta^\mu) \right]\ .
\end{equation}
To ensure stability in training, the target networks are updated slowly with the following soft update rules:
\begin{equation}
\begin{aligned}
	\theta^{Q'} & \leftarrow \tau \theta^Q + (1 - \tau) \theta^{Q'} \\
	\theta^{\mu'} & \leftarrow \tau \theta^\mu + (1 - \tau) \theta^{\mu'}
\end{aligned}
\ ,
\end{equation}
where $\tau$ is a small parameter. The reinforcement learning agent, through continuous interaction with the evolutionary process, collects more experience into its experience pool. This allows for continuous training to recommend more accurate operator portfolio schemes that are suitable for the current population states.

\section{Empirical Studies}

In this section, a series of experiments are conducted to validate the effectiveness of the proposed CMOEA-AOP in addressing CMOPs. Specifically, we verify the performance of the proposed CMOEA-AOP by comparing it to EMCMO \cite{qiao2022evolutionary}, Bico \cite{liu2021handling}, AGEMOEA-II \cite{panichella2022improved}, TSTI \cite{dong2022two}, and DRLOS \cite{ming2024constrained} on the constrained CF test suite \cite{zhang2008multiobjective}, LIR-CMOP test suite \cite{fan2019improved}, and DAS-CMOP test suite \cite{fan2020difficulty}. In addition, the effectiveness of the automated operator portfolio approach of CMOEA-AOP is verified by an ablation study. All the experiments are conducted on PlatEMO~\cite{tian2023practical}.

\subsection{Settings of Problems}

The number of objectives $M$ and the number of decision variables $D$ of each benchmark are set as follows. For the 10 CF problems, $M=2$, $D=10$ for CF1--7 and $M=3$ for CF8--10. For the 14 LIR--CMOP problems, $M=2$, $D=30$ for LIR-CMOP1--LIR-CMOP12 and $M=3$ for LIR-CMOP13 and LIR-CMOP14. For the 9 DAS-CMOP problems,  $M=2$, $D=30$ for DAS-CMOP1--6 and $M=3$ for DAS-CMOP7--9. For all constrained optimization problems, the quality of obtained feasible solutions is assessed using the inverted generational distance (IGD) \cite{bosman2003balance}, which involves sampling 10,000 reference points from each Pareto front \cite{tian2018sampling} to calculate the IGD value. The Wilcoxon rank sum test \cite{derrac2011practical}, with a significance level set at 0.05, is utilized to perform statistical analysis between each compared algorithm and the proposed CMOEA-AOP based on 30 independent runs on each test instance. In the results, `$+$' denotes that an algorithm performs significantly better than CMOEA-AOP, `$-$' indicates that an algorithm is significantly worse than CMOEA-AOP, and `$=$' suggests that an algorithm shows statistical similarity to CMOEA-AOP.

\subsection{Settings of Algorithms}

Five popular CMOEAs (i.e, EMCMO \cite{qiao2022evolutionary}, Bico \cite{liu2021handling}, AGEMOEA-II \cite{panichella2022improved}, TSTI \cite{dong2022two} and DRLOS \cite{ming2024constrained}) are selected as baselines. EMCMO \cite{qiao2022evolutionary} models a CMOP as a multitasking optimization problem and assigns two populations to optimize two tasks in a parallel way. Bico \cite{liu2021handling} maintains two populations: main and archive, where the main population is updated using constraint-domination and an NSGA-II variant to move it into the feasible region and guide it towards the Pareto front. AGEMOEA-II \cite{panichella2022improved} designs an accurate yet fast method to model the geometry of the non-dominated fronts. TSTI \cite{dong2022two} is a two-stage constrained evolutionary algorithm with different emphases on the three indicators. DRLOS \cite{ming2024constrained} is an evolutionary algorithm that embeds the adaptive operator selection into EMCMO using reinforcement learning methods. According to the taxonomy given in Section II-A, AGEMOEA-II belongs to the first category, TSTI belongs to the second category, Bico belongs the third category, and EMCMO and DRLOS belong to the fourth category.

All algorithms are configured with a uniform population size of 100 for all problem instances. They are then executed for a maximum of 100\,000 function evaluations on all problem instances. Additionally, all compared algorithms adopt parameter values recommended in their original literature to ensure fair comparisons. For EMCMO, the parameter $\beta$ to control the conversion of the evolutionary phase is set to 0.2. For TSTI, the parameters for calculation frank and for adjusting epsilon are set to 0.05 and 1.01, respectively. For DRLOS, the probability of simulated binary crossover is set to 1, the probability of polynomial mutation is set to $1/D$ ($D$ denotes the number of decision variables), the distribution index of both crossover and mutation is set to 20, and the parameters CR and F in differential evolution are set to 1 and 0.5, respectively. For the proposed CMOEA-AOP, the discount factor $\gamma$ is set to 0.98, and the batch size $BS$ is set to 32. 

\subsection{Experimental Results}

\begin{table*}[t!]
	\renewcommand{\arraystretch}{1.55}
	\centering
	\caption{IGD Values Obtained by EMCMO, BICO, AGEMOEA-II, DRLOS, and the Proposed CMOEA-AOP on the CF Benchmark Suite, LIR-CMOP Benchmark Suite, and DAS-CMOP Benchmark Suite. Best Result in Each Row Is Highlighted}
	\resizebox{\textwidth}{!}{
		\begin{tabular}{ccccccc}
			\toprule
			Problem&EMCMO&BiCo&AGEMOEA-II&TSTI&DRLOS&CMOEA-AOP\\
			\midrule
			\multirow{1}{*}{CF1}&5.8878e-3 (6.15e-4) $-$&1.6353e-2 (2.05e-3) $-$&2.1244e-2 (3.06e-3) $-$&2.6988e-2 (3.28e-3) $-$&2.3242e-3 (4.21e-4) $-$&\hl{1.8565e-3 (3.43e-4)}\\
			\multirow{1}{*}{CF2}&3.5223e-2 (1.33e-2) $-$&4.9712e-2 (1.55e-2) $-$&5.0383e-2 (1.93e-2) $-$&7.0321e-2 (2.01e-2) $-$&\hl{7.5237e-3 (3.09e-3) $\approx$}&8.1487e-3 (3.15e-3)\\
			\multirow{1}{*}{CF3}&2.1806e-1 (7.60e-2) $\approx$&2.3767e-1 (7.24e-2) $-$&3.0784e-1 (1.89e-1) $-$&3.3490e-1 (1.08e-1) $-$&2.7928e-1 (1.32e-1) $-$&\hl{1.8946e-1 (1.33e-1)}\\
			\multirow{1}{*}{CF4}&9.9130e-2 (3.04e-2) $-$&9.8463e-2 (3.58e-2) $-$&1.0766e-1 (3.35e-2) $-$&1.4982e-1 (3.89e-2) $-$&\hl{4.5235e-2 (8.80e-3) $+$}&5.2360e-2 (6.51e-3)\\
			\multirow{1}{*}{CF5}&2.7146e-1 (9.80e-2) $-$&3.0442e-1 (1.36e-1) $-$&3.2254e-1 (1.16e-1) $-$&4.0078e-1 (1.25e-1) $-$&2.2847e-1 (1.09e-1) $-$&\hl{1.8643e-1 (1.41e-1)}\\
			\multirow{1}{*}{CF6}&5.6069e-2 (2.43e-2) $-$&9.4654e-2 (4.28e-2) $-$&9.4154e-2 (3.35e-2) $-$&1.0310e-1 (2.49e-2) $-$&4.3410e-2 (2.53e-2) $\approx$&\hl{3.5271e-2 (6.68e-3)}\\
			\multirow{1}{*}{CF7}&2.2361e-1 (8.85e-2) $-$&2.5107e-1 (1.02e-1) $-$&3.2223e-1 (1.22e-1) $-$&3.3588e-1 (1.25e-1) $-$&1.8168e-1 (6.50e-2) $-$&\hl{1.3428e-1 (9.79e-2)}\\
			\multirow{1}{*}{CF8}&2.2052e-1 (9.72e-2) $-$&2.0679e-1 (7.78e-2) $\approx$&2.1640e-1 (1.15e-1) $\approx$&4.3384e-1 (2.67e-2) $-$&2.0446e-1 (3.26e-2) $-$&\hl{1.6928e-1 (2.63e-2)}\\
			\multirow{1}{*}{CF9}&1.0542e-1 (1.30e-2) $-$&\hl{9.6072e-2 (9.76e-3) $\approx$}&1.8777e-1 (8.12e-2) $-$&1.1183e-1 (1.81e-2) $-$&1.2348e-1 (1.85e-2) $-$&9.8709e-2 (9.33e-3)\\
			\multirow{1}{*}{CF10}&3.6302e-1 (7.43e-2) $\approx$&4.8058e-1 (3.06e-1) $\approx$&NaN (NaN)&4.9545e-1 (1.40e-1) $-$&3.7665e-1 (7.97e-2) $\approx$&\hl{3.4425e-1 (8.56e-2)}\\
			\hline
			\multirow{1}{*}{LIR-CMOP1}&2.5495e-1 (5.38e-2) $-$&2.2050e-1 (1.51e-2) $\approx$&3.0052e-1 (3.22e-2) $-$&2.2497e-1 (2.17e-2) $-$&2.0550e-1 (4.89e-2) $\approx$&\hl{1.8383e-1 (6.45e-2)}\\
			\multirow{1}{*}{LIR-CMOP2}&2.4457e-1 (3.28e-2) $-$&1.8610e-1 (2.26e-2) $-$&2.5828e-1 (2.50e-2) $-$&1.9268e-1 (1.84e-2) $-$&1.8352e-1 (4.38e-2) $-$&\hl{1.3838e-1 (5.34e-2)}\\
			\multirow{1}{*}{LIR-CMOP3}&2.5254e-1 (5.59e-2) $-$&2.1764e-1 (2.66e-2) $\approx$&3.0987e-1 (3.02e-2) $-$&2.2308e-1 (2.54e-2) $-$&2.2637e-1 (6.24e-2) $-$&\hl{1.9710e-1 (4.99e-2)}\\
			\multirow{1}{*}{LIR-CMOP4}&2.7398e-1 (3.39e-2) $-$&2.1737e-1 (2.43e-2) $\approx$&2.9239e-1 (3.00e-2) $-$&2.2211e-1 (2.52e-2) $\approx$&2.3637e-1 (4.29e-2) $-$&\hl{1.9150e-1 (6.19e-2)}\\
			\multirow{1}{*}{LIR-CMOP5}&2.8745e-1 (5.99e-2) $-$&1.2247e+0 (6.14e-3) $-$&1.2187e+0 (7.44e-3) $-$&9.5034e-1 (4.14e-1) $-$&\hl{1.7759e-2 (5.40e-3) $+$}&2.9712e-2 (3.01e-2)\\
			\multirow{1}{*}{LIR-CMOP6}&2.9387e-1 (7.97e-2) $-$&1.3453e+0 (1.79e-4) $-$&1.3448e+0 (2.62e-4) $-$&1.0272e+0 (4.29e-1) $-$&1.4260e-2 (4.01e-3) $\approx$&\hl{1.4253e-2 (6.10e-3)}\\
			\multirow{1}{*}{LIR-CMOP7}&1.0681e-1 (2.24e-2) $-$&6.0556e-1 (7.17e-1) $-$&7.1201e-1 (7.50e-1) $-$&1.4822e-1 (3.77e-2) $-$&1.6363e-2 (2.18e-2) $-$&\hl{9.0487e-3 (1.68e-3)}\\
			\multirow{1}{*}{LIR-CMOP8}&1.6762e-1 (5.63e-2) $-$&1.3073e+0 (6.33e-1) $-$&1.2070e+0 (6.81e-1) $-$&3.9529e-1 (4.40e-1) $-$&8.7959e-3 (1.64e-3) $\approx$&\hl{8.3729e-3 (5.78e-4)}\\
			\multirow{1}{*}{LIR-CMOP9}&4.7977e-1 (1.08e-1) $-$&9.5571e-1 (1.23e-1) $-$&9.7187e-1 (1.12e-1) $-$&5.2817e-1 (6.62e-2) $-$&2.4138e-1 (3.95e-2) $-$&\hl{2.2759e-1 (4.09e-2)}\\
			\multirow{1}{*}{LIR-CMOP10}&1.8859e-1 (8.80e-2) $-$&9.6708e-1 (5.75e-2) $-$&8.5218e-1 (1.15e-1) $-$&5.4914e-1 (1.81e-1) $-$&5.6143e-2 (4.07e-2) $-$&\hl{3.8295e-2 (2.96e-2)}\\
			\multirow{1}{*}{LIR-CMOP11}&6.2460e-2 (3.76e-2) $-$&7.1368e-1 (1.82e-1) $-$&7.5691e-1 (9.50e-2) $-$&5.2726e-1 (1.60e-1) $-$&2.5776e-2 (2.23e-2) $\approx$&\hl{2.0293e-2 (1.53e-2)}\\
			\multirow{1}{*}{LIR-CMOP12}&2.2556e-1 (7.99e-2) $-$&6.2927e-1 (2.22e-1) $-$&8.0366e-1 (1.70e-1) $-$&3.6104e-1 (7.10e-2) $-$&9.7681e-2 (4.41e-2) $-$&\hl{8.0086e-2 (2.47e-2)}\\
			\multirow{1}{*}{LIR-CMOP13}&\hl{9.0818e-2 (7.07e-4) $+$}&1.3183e+0 (1.61e-3) $-$&1.3053e+0 (1.28e-3) $-$&1.1574e+0 (4.11e-1) $-$&1.0399e-1 (2.20e-3) $-$&9.3102e-2 (7.76e-4)\\
			\multirow{1}{*}{LIR-CMOP14}&\hl{9.5810e-2 (5.83e-4) $+$}&1.2750e+0 (2.13e-3) $-$&1.2618e+0 (9.95e-4) $-$&1.0390e+0 (4.73e-1) $-$&9.8204e-2 (1.17e-3) $-$&9.6579e-2 (8.81e-4)\\
			\hline
			\multirow{1}{*}{DAS-CMOP1}&7.4530e-1 (3.33e-2) $-$&7.5955e-1 (2.67e-2) $-$&7.4741e-1 (3.49e-2) $-$&6.6527e-1 (6.52e-2) $-$&3.3048e-1 (3.50e-1) $-$&\hl{4.5018e-3 (1.10e-3)}\\
			\multirow{1}{*}{DAS-CMOP2}&2.3726e-1 (1.90e-2) $-$&2.3850e-1 (2.00e-2) $-$&2.6862e-1 (3.28e-2) $-$&2.3616e-1 (5.12e-2) $-$&5.5651e-3 (3.08e-4) $-$&\hl{5.2974e-3 (3.17e-4)}\\
			\multirow{1}{*}{DAS-CMOP3}&3.4143e-1 (4.03e-2) $-$&3.2092e-1 (6.05e-2) $-$&3.6482e-1 (4.09e-2) $-$&3.2547e-1 (5.03e-2) $-$&2.2111e-2 (4.54e-3) $-$&\hl{2.0294e-2 (7.24e-4)}\\
			\multirow{1}{*}{DAS-CMOP4}&\hl{2.7847e-3 (1.73e-3) $+$}&1.0256e-1 (1.33e-1) $\approx$&1.7470e-1 (1.47e-1) $-$&1.0524e-1 (1.31e-1) $\approx$&1.6952e-2 (4.89e-2) $\approx$&6.8261e-2 (1.37e-1)\\
			\multirow{1}{*}{DAS-CMOP5}&\hl{3.1285e-3 (1.42e-3) $+$}&8.8530e-2 (1.66e-1) $-$&3.1640e-1 (2.36e-1) $-$&2.8626e-2 (9.83e-2) $+$&7.6055e-3 (1.34e-2) $\approx$&3.1157e-2 (1.00e-1)\\
			\multirow{1}{*}{DAS-CMOP6}&\hl{3.8721e-2 (3.82e-2) $\approx$}&1.5110e-1 (1.95e-1) $\approx$&5.0734e-1 (1.03e-1) $-$&3.9263e-1 (1.89e-1) $-$&5.0920e-2 (9.87e-2) $\approx$&1.9644e-1 (2.62e-1)\\
			\multirow{1}{*}{DAS-CMOP7}&3.1111e-2 (1.24e-3) $-$&3.1962e-2 (1.52e-3) $-$&3.1285e-2 (6.33e-4) $-$&3.1414e-2 (1.79e-3) $-$&6.8528e-2 (8.04e-2) $-$&\hl{3.0523e-2 (6.86e-4)}\\
			\multirow{1}{*}{DAS-CMOP8}&3.9996e-2 (8.60e-4) $-$&4.0815e-2 (1.40e-3) $-$&4.6516e-2 (1.79e-3) $-$&4.0091e-2 (8.93e-4) $-$&9.9003e-2 (9.44e-2) $-$&\hl{3.9156e-2 (1.08e-3)}\\
			\multirow{1}{*}{DAS-CMOP9}&3.5359e-1 (6.10e-2) $-$&3.5867e-1 (4.91e-2) $-$&3.9066e-1 (9.56e-2) $-$&2.8630e-1 (7.18e-2) $-$&\hl{4.1910e-2 (1.24e-3) $+$}&4.5292e-2 (4.01e-3)\\
			\hline
			\multicolumn{1}{c}{$+/-/\approx$}&4/26/3&0/25/8&0/31/1&1/30/2&3/20/10&\\
			\bottomrule
		\end{tabular}
	}
	\label{tab:benchmark}
\end{table*}

\begin{figure*}[!t]
	\centering
	\subfigure{\includegraphics[width=0.75\linewidth]{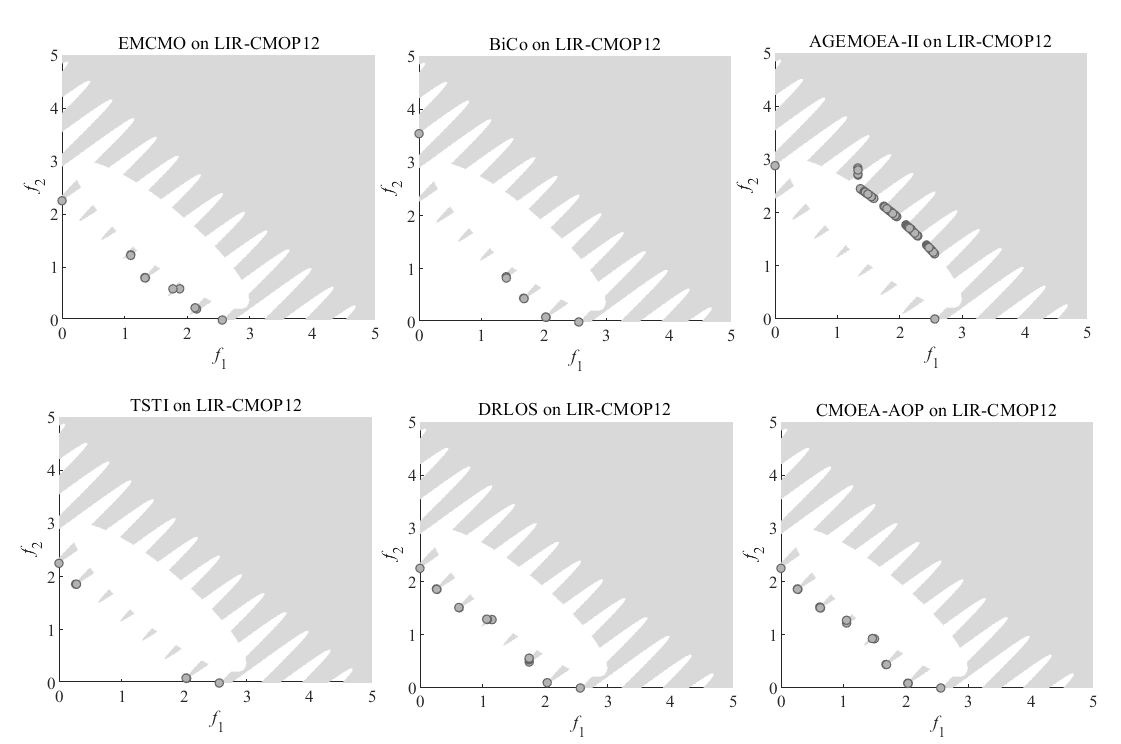}}\\
	\caption{Populations with the median IGD obtained by EMCMO, Bico, AGEMOEA-II, DRLOS, and the proposed CMOEA-AOP on LIR-CMOP12.}
	\label{fig:objLIR-CMOP12}
\end{figure*}


\begin{figure*}[!t]
	\centering
	\subfigure{\includegraphics[width=0.75\linewidth]{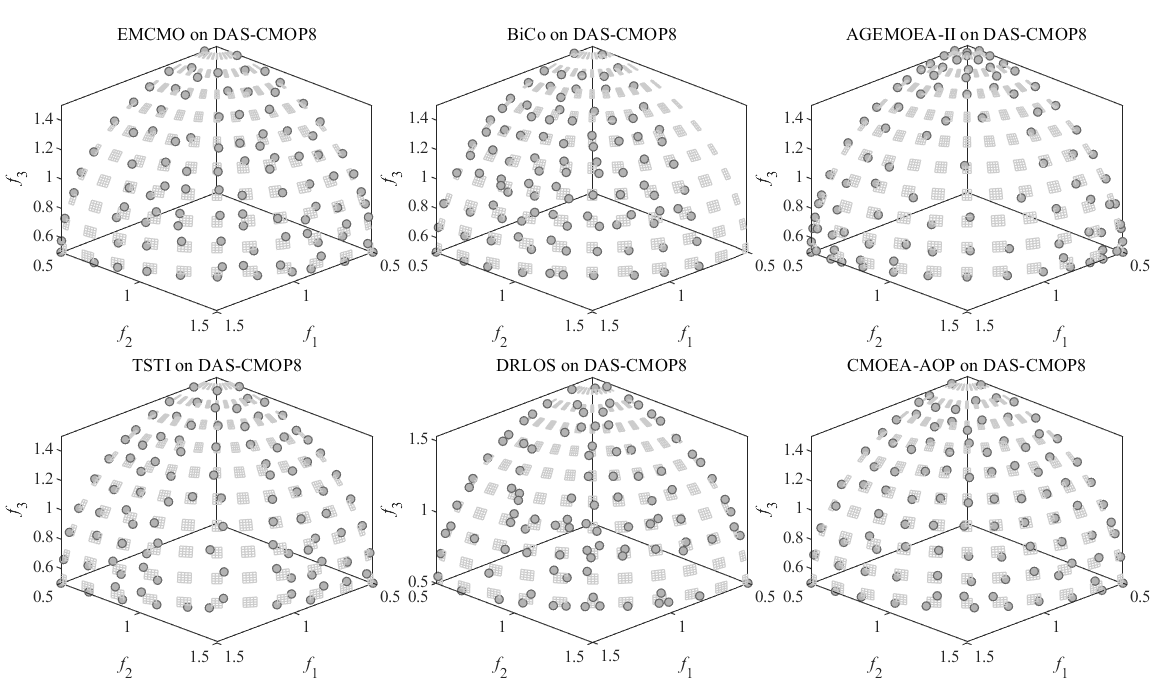}}\\
	\caption{\XG{Populations with the median IGD obtained by EMCMO, Bico, AGEMOEA-II, DRLOS, and the proposed CMOEA-AOP on DAS-CMOP8.}}
	\label{fig:objDAS-CMOP8}
\end{figure*}

\begin{figure*}[!t]
	\centering
	\subfigure{\includegraphics[width=0.85\linewidth]{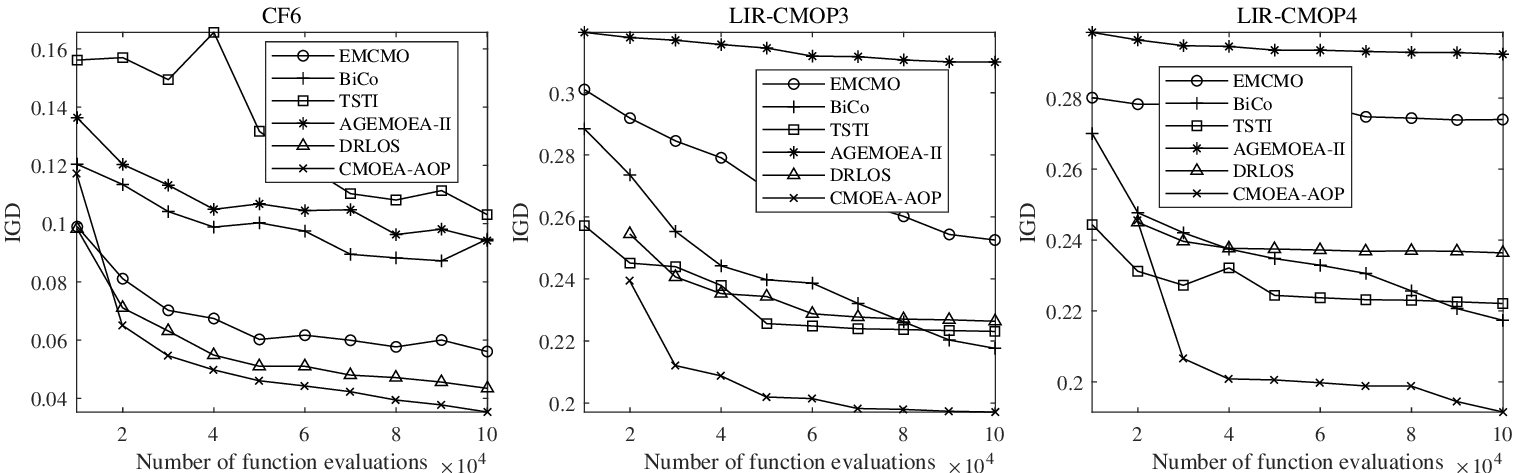}}\\
	\caption{Convergence profiles of mean IGD values by EMCMO, Bico, AGEMOEA-II, DRLOS, and the proposed CMOEA-AOP on CF6, LIR-CMOP3 and LIR-CMOP4.}
	\label{fig:MOPConvergence}
\end{figure*}

The optimization results of the proposed CMOEA-AOP and five comparative algorithms on the CMOPs are presented in Table \ref{tab:benchmark}. It can be observed that CMOEA-AOP performs the best on most problem instances. Specifically, out of 33 test instances in total, CMOEA-AOP achieves the best results in 23 instances. In terms of statistical testing, CMOEA-AOP significantly outperforms EMCMO, Bico, AGEMOEA-II, TSTI and DRLOS on 26, 25, 31, 30, and 20 test instances, respectively. Therefore, the proposed CMOEA-AOP outperforms existing constrained MOEAs. In addition, Fig. \ref{fig:objLIR-CMOP12} plots the populations with the median IGD obtained by the six CMOEAs on LIR-CMOP12. The feasible region of LIR-CMOP12 is not only narrow but separate from each other; that is, it is difficult to process with existing CMOEAs. As shown in Fig. \ref{fig:objLIR-CMOP12}, obviously, AGEMOEA-II can only find a set of local optimal solutions located within the feasible region, while TSTI can only find a portion of solutions located on the edge of the feasible region. In contrast, EMCMO, Bico, and DRLOS can find some solutions that approach the global optimum, while the proposed CMOEA-AOP can find a large number of evenly distributed feasible solutions. It is evident that, although using adaptive operators, compared to DRLOS, which can only learn and recommend a single operator in a single iteration, the proposed CMOEA-AOP has stronger robustness and conducts more detailed searches. \XG{Fig. \ref{fig:objDAS-CMOP8} plots the population with the median IGD obtained by the six CMOEAs on DAS-CMOP8. It is evident that the proposed algorithm still maintains an advantage compared to its counterparts. It not only identifies feasible solutions but also exhibits uniform distribution across them.} For a further distinction between the performance of the six CMOEAs, Fig. \ref{fig:MOPConvergence} plots the convergence profiles of IGD values obtained by these algorithms on the same CMOPs. It can be clearly seen that the proposed CMOEA-AOP converges faster than  EMCMO, Bico, AGEMOEA-II, TSTI and DRLOS; in particular, the populations obtained by CMOEA-AOP with just 50\,000 function evaluations are competitive to the populations obtained by EMCMO, Bico, AGEMOEA-II, TSTI and DRLOS with 100\,000 function evaluations on CF6, LIR-CMOP3 and LIR-CMOP4.

It is worth noting that CMOEA-AOP can be embedded into most existing CMOEAs. In our experiments, we embedded it into EMCMO. As shown in Table \ref{tab:benchmark}, although the proposed algorithm adopts strategies consistent with EMCMO in constraint handling mechanisms, its performance far surpasses that of EMCMO. This further demonstrates the advantages of our proposed automated operator portfolio approach. In the next subsection, we will provide a more comprehensive set of ablation experiments to further illustrate the role of the proposed automated operator portfolio.

\subsection{Ablation Studies}

\begin{table}[t!]
	\renewcommand{\arraystretch}{1.3}
	\centering
	\caption{IGD Values Obtained by CMOEA-AOP and Its Three Variants on the CF Benchmark Suite, LIR-CMOP Benchmark Suite, and DAS-CMOP Benchmark Suite. Best Result in Each Row Is Highlighted}
	\scalebox{0.75}{
		\begin{tabular}{cccccc}
			\toprule
			Problem&CMOEA-AOP1&CMOEA-AOP2&CMOEA-AOP3&CMOEA-AOP\\
			\midrule
			\multirow{1}{*}{CF1}&3.9110e-3 $-$&\hl{1.6821e-3 $\approx$}&4.9569e-3 $-$&1.8565e-3\\
			\multirow{1}{*}{CF2}&7.1605e-2 $-$&9.0358e-3 $\approx$&1.5847e-2 $-$&\hl{8.1487e-3}\\
			\multirow{1}{*}{CF3}&2.4892e-1 $-$&2.5020e-1 $-$&\hl{1.3385e-1 $\approx$}&1.8946e-1\\
			\multirow{1}{*}{CF4}&1.4285e-1 $-$&5.3927e-2 $\approx$&5.9382e-2 $-$&\hl{5.2360e-2}\\
			\multirow{1}{*}{CF5}&2.8200e-1 $-$&2.8397e-1 $-$&2.0711e-1 $\approx$&\hl{1.8643e-1}\\
			\multirow{1}{*}{CF6}&6.9170e-2 $-$&3.8727e-2 $-$&4.6500e-2 $-$&\hl{3.5271e-2}\\
			\multirow{1}{*}{CF7}&3.1681e-1 $-$&1.7021e-1 $-$&1.6963e-1 $-$&\hl{1.3428e-1}\\
			\multirow{1}{*}{CF8}&\hl{1.2909e-1 $+$}&2.6485e-1 $-$&1.3212e-1 $+$&1.6928e-1\\
			\multirow{1}{*}{CF9}&\hl{6.9542e-2 $+$}&1.3489e-1 $-$&7.6262e-2 $+$&9.8709e-2\\
			\multirow{1}{*}{CF10}&\hl{2.5232e-1 $+$}&6.6155e-1 $-$&2.6334e-1 $+$&3.4425e-1\\
			\hline
			\multirow{1}{*}{LIR-CMOP1}&3.5160e-1 $-$&\hl{1.1718e-1 $+$}&2.1241e-1 $\approx$&1.8383e-1\\
			\multirow{1}{*}{LIR-CMOP2}&2.8224e-1 $-$&\hl{8.2993e-2 $+$}&1.0005e-1 $+$&1.3838e-1\\
			\multirow{1}{*}{LIR-CMOP3}&3.3363e-1 $-$&\hl{1.2726e-1 $+$}&2.0080e-1 $\approx$&1.9710e-1\\
			\multirow{1}{*}{LIR-CMOP4}&3.1351e-1 $-$&\hl{1.2285e-1 $+$}&1.9359e-1 $\approx$&1.9150e-1\\
			\multirow{1}{*}{LIR-CMOP5}&2.0709e-1 $-$&2.1215e-1 $-$&3.6296e-2 $\approx$&\hl{2.9712e-2}\\
			\multirow{1}{*}{LIR-CMOP6}&1.9571e-1 $-$&5.9811e-2 $-$&7.7148e-2 $-$&\hl{1.4253e-2}\\
			\multirow{1}{*}{LIR-CMOP7}&9.0441e-2 $-$&1.8475e-2 $-$&1.2934e-2 $-$&\hl{9.0487e-3}\\
			\multirow{1}{*}{LIR-CMOP8}&1.3749e-1 $-$&9.8532e-3 $-$&9.2371e-3 $-$&\hl{8.3729e-3}\\
			\multirow{1}{*}{LIR-CMOP9}&2.9734e-1 $-$&\hl{2.1451e-1 $\approx$}&4.4647e-1 $-$&2.2759e-1\\
			\multirow{1}{*}{LIR-CMOP10}&1.4405e-1 $-$&\hl{3.8094e-2 $\approx$}&2.1615e-1 $-$&3.8295e-2\\
			\multirow{1}{*}{LIR-CMOP11}&3.5810e-2 $\approx$&\hl{1.6296e-2 $\approx$}&8.8751e-2 $-$&2.0293e-2\\
			\multirow{1}{*}{LIR-CMOP12}&8.0546e-2 $\approx$&\hl{7.4248e-2 $\approx$}&1.7983e-1 $-$&8.0086e-2\\
			\multirow{1}{*}{LIR-CMOP13}&2.0465e-1 $-$&1.1976e-1 $-$&9.5196e-2 $-$&\hl{9.3102e-2}\\
			\multirow{1}{*}{LIR-CMOP14}&1.8277e-1 $-$&1.0263e-1 $-$&\hl{9.6545e-2 $\approx$}&9.6579e-2\\
			\hline
			\multirow{1}{*}{DAS-CMOP1}&6.9774e-1 $-$&\hl{3.9865e-3 $\approx$}&6.3593e-3 $-$&4.5018e-3\\
			\multirow{1}{*}{DAS-CMOP2}&2.3075e-1 $-$&\hl{5.2080e-3 $\approx$}&5.6341e-3 $-$&5.2974e-3\\
			\multirow{1}{*}{DAS-CMOP3}&3.0597e-1 $-$&2.0744e-2 $\approx$&2.1079e-2 $-$&\hl{2.0294e-2}\\
			\multirow{1}{*}{DAS-CMOP4}&2.2154e-1 $-$&\hl{4.8629e-2 $\approx$}&NaN&6.8261e-2\\
			\multirow{1}{*}{DAS-CMOP5}&1.5576e-1 $-$&9.1851e-2 $-$&NaN&\hl{3.1157e-2}\\
			\multirow{1}{*}{DAS-CMOP6}&2.1899e-1 $-$&\hl{5.5946e-2 $\approx$}&NaN&1.9644e-1\\
			\multirow{1}{*}{DAS-CMOP7}&2.9568e-1 $-$&1.8798e-1 $-$&NaN&\hl{3.0523e-2}\\
			\multirow{1}{*}{DAS-CMOP8}&3.6835e-1 $-$&1.5872e-1 $-$&5.8937e-1 $\approx$&\hl{3.9156e-2}\\
			\multirow{1}{*}{DAS-CMOP9}&4.6723e-1 $-$&4.3900e-2 $\approx$&\hl{4.3498e-2 $\approx$}&4.5292e-2\\
			\hline
			\multicolumn{1}{c}{$+/-/\approx$}&3/28/2&4/16/13&4/16/9&\\
			\bottomrule
		\end{tabular}
	}
	\label{tab:abla}
\end{table}

This section performs an ablation study to verify the effectiveness of the core components of the proposed CMOEA-AOP, where CMOEA-AOP is compared to its  variants with a single operator on the CF Benchmark Suite, LIR-CMOP Benchmark Suite and DAS-CMOP Benchmark Suite, so that the influence brought by the difference between other strategies can be totally eliminated. Table \ref{tab:abla} presents the statistical results of CMOEA-AOP and its three variants, where CMOEA-AOP1, CMOEA-AOP2, and CMOEA-AOP3 represent using only genetic operators, DE/rand/1, and DE/best/1, respectively. Although the proposed CMOEA-AOP achieved the best results on only 14 instances compared to its five variants, the statistical results indicate that CMOEA-AOP outperforms the other algorithms. To be specific, in terms of statistical testing, CMOEA-AOP significantly outperforms CMOEA-AOP1, CMOEA-AOP2, and CMOEA-AOP3 on 28, 16, and 16 test instances, respectively. This demonstrates that the variants of CMOEA-AOP, which employs a single operator, may achieve significant advantages on a small number of problems, but it does not possess generalization capabilities and may be at a disadvantage on most other problems. Therefore, the proposed reinforcement learning-assisted operator portfolio scheme is not only effective but also exhibits extremely stable performance for handling CMOPs.

\section{Conclusion}

To effectively handle CMOPs, operator adaptation holds significant promise. However, existing operator adaptation approaches can only recommend a single operator in each iteration, making it challenging for evolutionary algorithms to strike a good balance between exploration and exploitation. To enhance the search capability of algorithms by introducing multiple search paradigms in a single iteration, we have developed a CMOEA embedded with a deep reinforcement learning-assisted automated operator portfolio approach. The proposed algorithm trains a reinforcement learning agent by using different portfolio ratios of operators as actions and extracts optimization and constraint-related features from the current population during the evolutionary process. Consequently, the agent can recommend the best operator portfolio scheme for the current population, thereby rapidly approaching the constrained Pareto front. Experimental results demonstrate significant advantages of the proposed algorithm on three sets of benchmark functions.

\XG{In future research, we aim to further extend the proposed automated operator portfolio to solve unconstrained MOPs \cite{tian2021evolutionary}, thereby enhancing its applicability in real-world scenarios. Additionally, since the proposed algorithm utilizes the overall constraint violation of each solution as its feasibility feature, it is promising to extend it to consider each constraint function separately to further improve its performance. Finally, considering that training neural networks using reinforcement learning methods requires considerable time resources, we intend to explore the use of pre-trained large language models \cite{kasneci2023chatgpt, zhao2023survey} for automated operator portfolio as a future research direction.}


\ifCLASSOPTIONcaptionsoff
  \newpage
\fi
\bibliographystyle{IEEEtran}
\bibliography{references}

\begin{IEEEbiography}[{\includegraphics[width=1in,height=1.2in,clip,keepaspectratio]{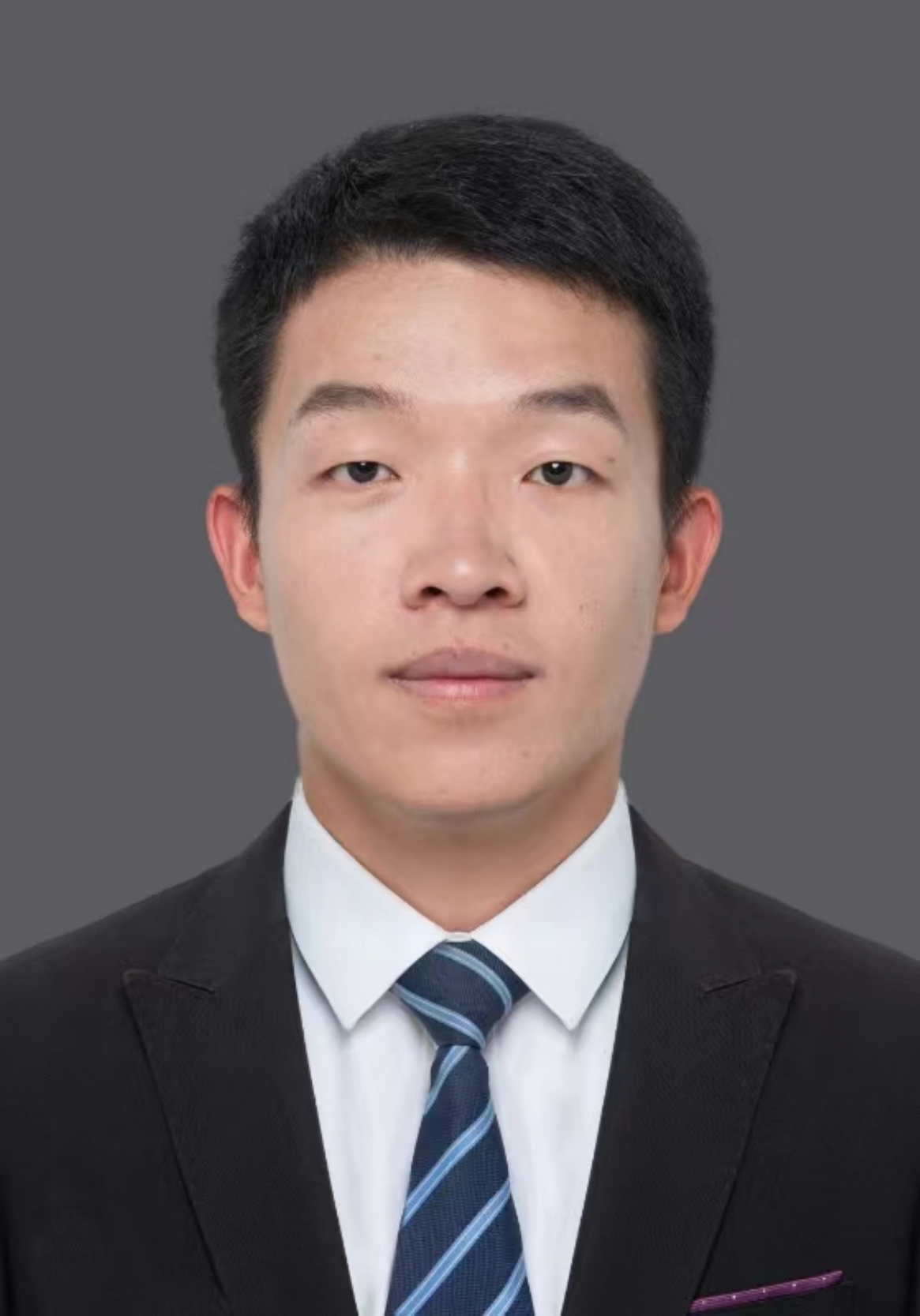}}]{Shuai Shao}
	received the B.Sc. degree from Anhui University, Hefei, China, in 2021, where he is currently pursuing the Ph.D. degree with the School of Computer Science and Technology.
	
	His current research interests include evolutionary multiobjective optimization, deep reinforcement learning, and automated machine learning. He is the recipient of the 18th International Conference on Bio-inspired Computing: Theories and Applications Best Paper Award.
\end{IEEEbiography}

\begin{IEEEbiography}[{\includegraphics[width=1in,height=1.2in,clip,keepaspectratio]{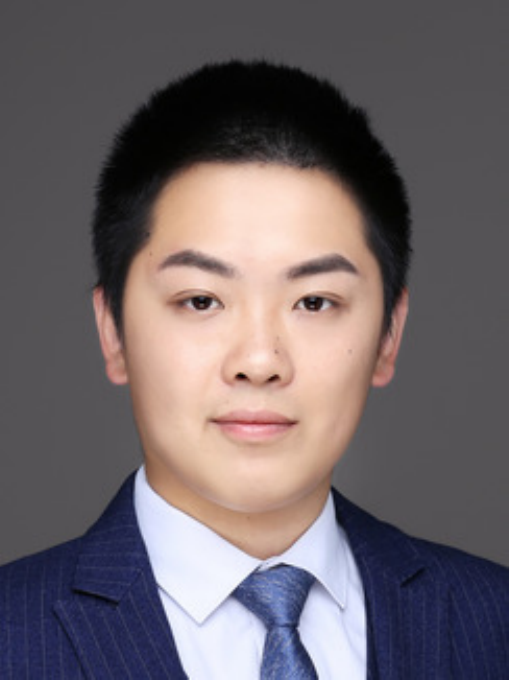}}]{Ye Tian (Senior Member, IEEE)}
	received the B.Sc., M.Sc., and Ph.D. degrees from Anhui University, Hefei, China, in 2012, 2015, and 2018, respectively.
	
	He is currently a Professor with the School of Computer Science and Technology, Anhui University, Hefei, China. His current research interests include evolutionary computation and its applications. He is the recipient of the 2018, 2021, and 2024 IEEE Transactions on Evolutionary Computation Outstanding Paper Award, the 2020 IEEE Computational Intelligence Magazine Outstanding Paper Award, and the 2022 IEEE Computational Intelligence Society Outstanding PhD Dissertation Award.
\end{IEEEbiography}

\begin{IEEEbiography}[{\includegraphics[width=1in,height=1.2in,clip,keepaspectratio]{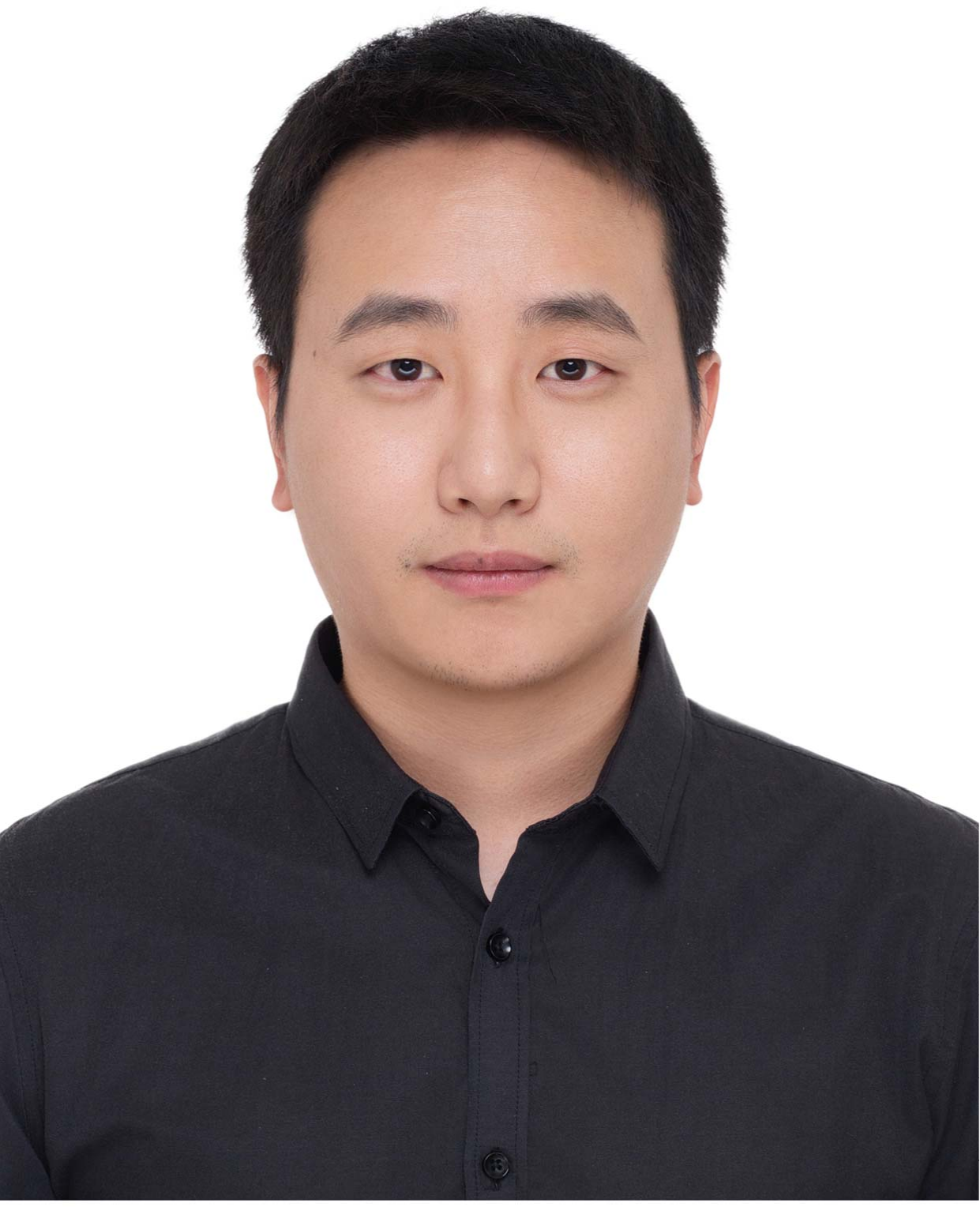}}]{Shangshang Yang}
	received the B.Sc. and Ph.D. degrees from Anhui University, Hefei, China, in 2017 and 2022, respectively.
	
	He is currently a postdoctoral researcher at the School of Computer Science and Technology, Anhui University, Hefei, China. His current research interests include evolutionary multi-objective optimization, automated machine learning, and intelligent education. He is the recipient of the 2023 International Conference on Data-driven Optimization of Complex Systems Best Paper Award. He received the Postdoctoral Fellowship Program (Grade B) and was awarded a Research Performance Assessment Grant (Second Grade) from the China Postdoctoral Science Foundation. He served as a reviewer/PC member for multiple top-tier conferences, including NeurIPS, ICML, ICLR, AAAI, and IJCAI.
\end{IEEEbiography}

\begin{IEEEbiography}[{\includegraphics[width=1in,height=1.2in,clip,keepaspectratio]{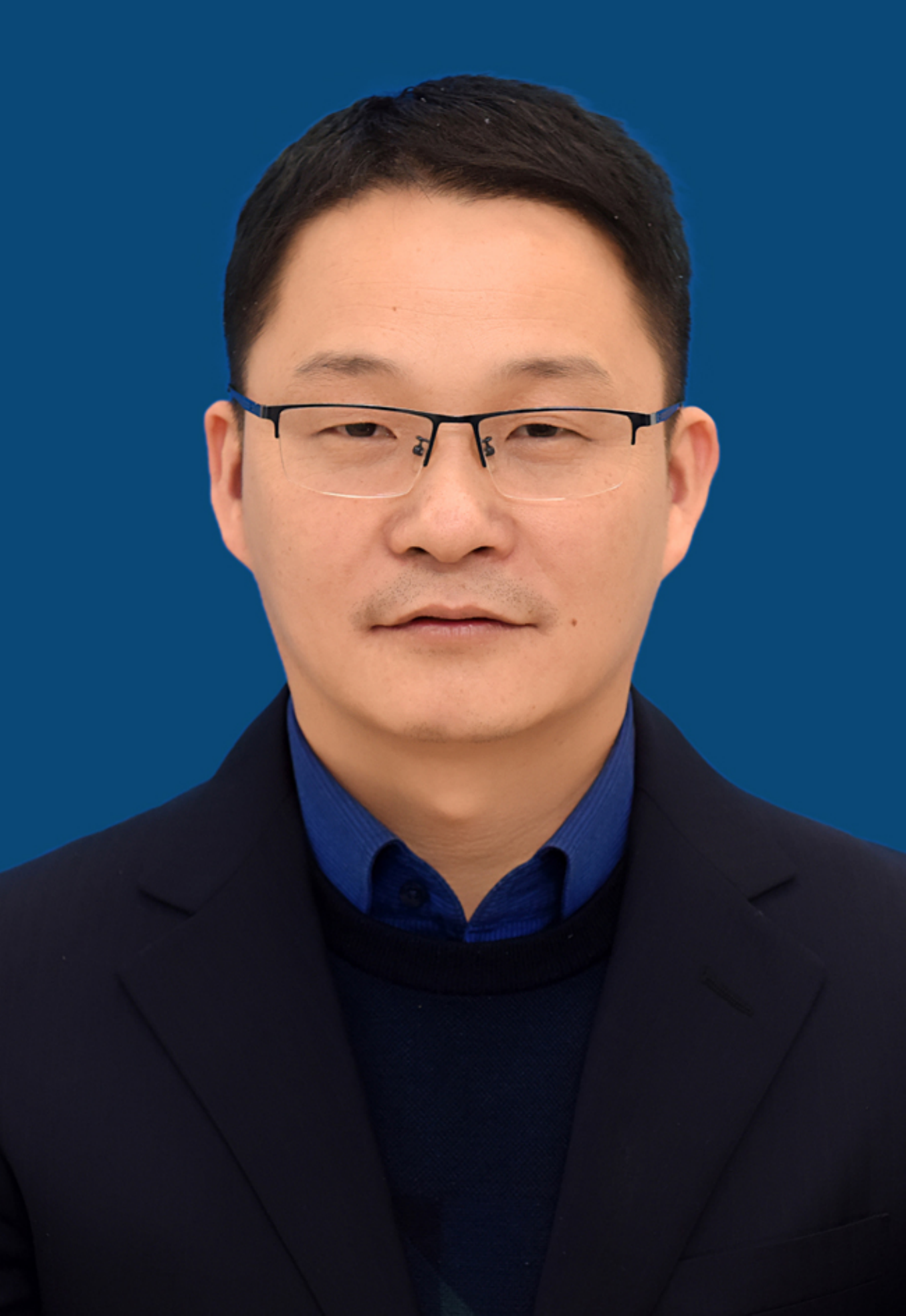}}]{Xingyi Zhang (Fellow, IEEE)} received the B.Sc. degree from Fuyang Normal College, Fuyang, China, in 2003, and the M.Sc. and Ph.D. degrees from Huazhong University of Science and Technology, Wuhan, China, in 2006 and 2009, respectively.
	
	He is currently a Professor with the School of Computer Science and Technology, Anhui University, Hefei, China. His current research interests include unconventional models and algorithms of computation, evolutionary multi-objective optimization, and logistic scheduling. He is the recipient of the 2018, 2021, and 2024 IEEE Transactions on Evolutionary Computation Outstanding Paper Award and the 2020 IEEE Computational Intelligence Magazine Outstanding Paper Award.
\end{IEEEbiography}

\end{document}